\documentclass[10pt, conference]{IEEEtran}
\IEEEoverridecommandlockouts
\usepackage{cite}
\usepackage{amsmath, amssymb, amsfonts}
\usepackage{algorithmic}
\usepackage{graphicx}
\usepackage{textcomp}
\usepackage{xcolor}
\usepackage{booktabs}
\usepackage{multirow}
\usepackage{array}
\usepackage{makecell} 
\usepackage{longtable}
\usepackage{caption} 
\usepackage{subcaption}
\usepackage{tikz}
\usepackage{pgfplots}
\pgfplotsset{compat=1.18}
\usepackage{placeins} 
\usepackage{dblfloatfix} 
\usepackage{url}
\usepackage{hyperref}
\usepackage{tabularx}

\IEEEaftertitletext{\vspace{-2\baselineskip}}

\def\BibTeX{{\rm B\kern-.05em{\sc i\kern-.025em b}\kern-.08em T\kern-.1667em\lower.7ex\hbox{E}\kern-.125emX}}

\IEEEoverridecommandlockouts
\IEEEpubid{\makebox[\columnwidth]{979-8-3315-8654-6/25/\$31.00~\copyright2025 IEEE \hfill} \hspace{\columnsep}\makebox[\columnwidth]{ }}

\begin{document}
    \title{Performance Evaluation of Deep Learning for Tree Branch Segmentation
    in Autonomous Forestry Systems}
    \author{\IEEEauthorblockN{Yida Lin, Bing Xue, Mengjie Zhang} \IEEEauthorblockA{\small \textit{Centre for Data Science and Artificial Intelligence} \\ \textit{Victoria University of Wellington, Wellington, New Zealand}\\ linyida\texttt{@}myvuw.ac.nz, bing.xue\texttt{@}vuw.ac.nz, mengjie.zhang\texttt{@}vuw.ac.nz}
    \and \IEEEauthorblockN{Sam Schofield, Richard Green} \IEEEauthorblockA{\small \textit{Department of Computer Science and Software Engineering} \\ \textit{University of Canterbury, Canterbury, New Zealand}\\ sam.schofield\texttt{@}canterbury.ac.nz, richard.green\texttt{@}canterbury.ac.nz}
    }
    \maketitle
    \IEEEpubidadjcol
    \vspace{-1.5em}

    \begin{abstract}
        UAV-based autonomous forestry operations require rapid and precise tree branch
        segmentation for safe navigation and automated pruning across varying pixel
        resolutions and operational conditions. We evaluate different deep
        learning methods at three resolutions (256×256, 512×512, 1024×1024)
        using the Urban Street Tree Dataset, employing standard metrics (IoU, Dice)
        and specialized measures including Thin Structure IoU (TS-IoU) and
        Connectivity Preservation Rate (CPR). Among 22 configurations tested, U-Net
        with MiT-B4 backbone achieves strong performance at 256×256. At 512×512,
        MiT-B4 leads in IoU, Dice, TS-IoU, and Boundary-F1. At 1024×1024, U-Net+MiT-B3
        shows the best validation performance for IoU/Dice and precision, while
        U-Net++ excels in boundary quality. PSPNet provides the most efficient option
        (2.36/9.43/37.74 GFLOPs) with 25.7/19.6/11.8 percentage point IoU reductions
        compared to top performers at respective resolutions. These results
        establish multi-resolution benchmarks for accuracy–efficiency trade-offs
        in embedded forestry systems. Implementation is available at \url{https://github.com/BennyLinntu/Performance_Tree_Branch_Segmentation}.
    \end{abstract}

    \vspace{-1.5em}
    \section{Introduction}

    Autonomous forestry operations have potential to improve worker safety and
    operational efficiency in tree maintenance tasks. Manual pruning operations require
    workers to operate at heights, often near power lines, creating occupational
    risks. Unmanned Aerial Vehicles (UAVs) equipped with computer vision systems
    offer an alternative approach through automated branch identification and
    pruning operations~\cite{Steininger2025TimberVision}.

\begin{figure}[htb]
        \centering
        \begin{subfigure}
            [t]{0.22\textwidth}
            \centering
            \includegraphics[width=\linewidth]{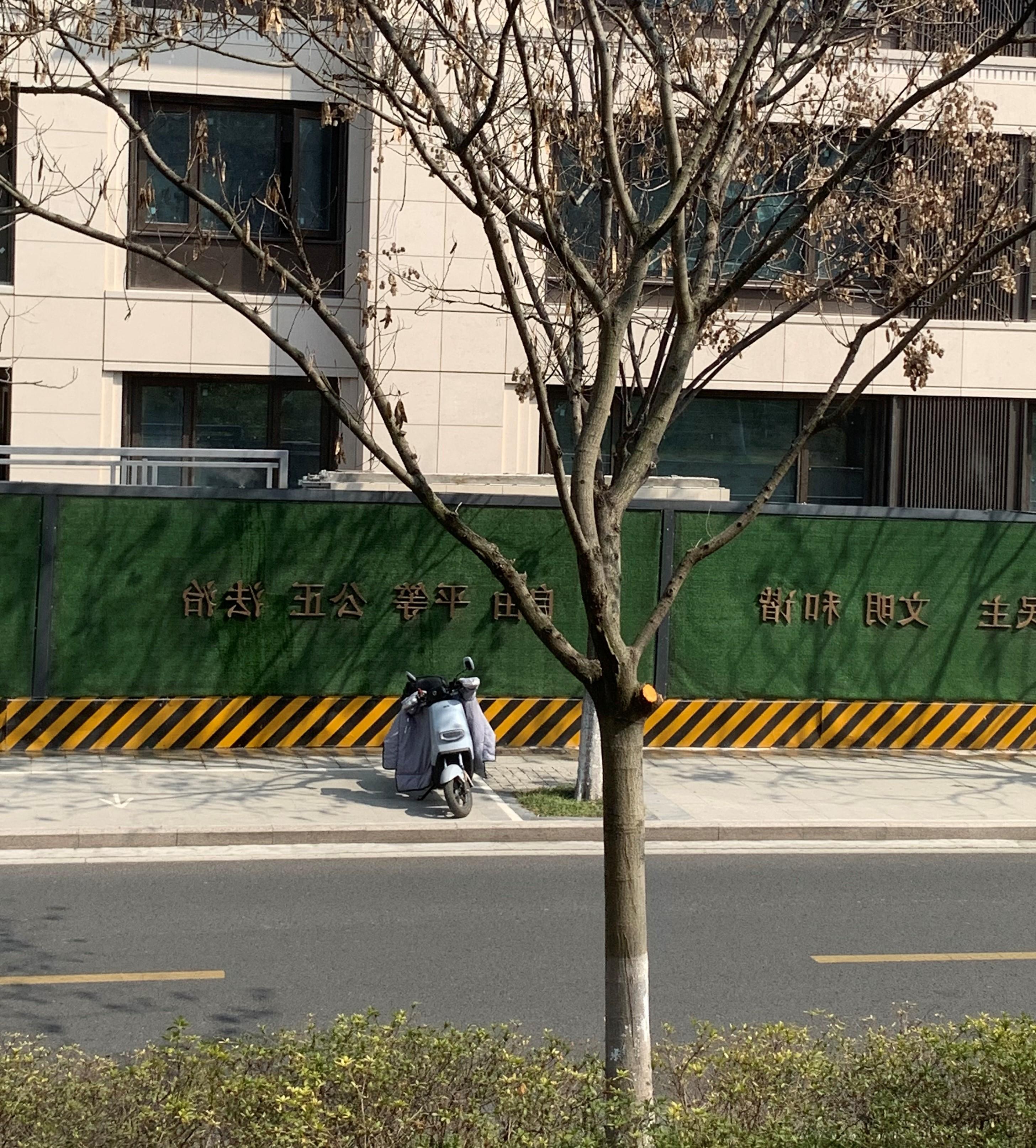}
            \caption{tree RGB image}
        \end{subfigure}\hfill
        \begin{subfigure}
            [t]{0.22\textwidth}
            \centering
            \includegraphics[width=\linewidth]{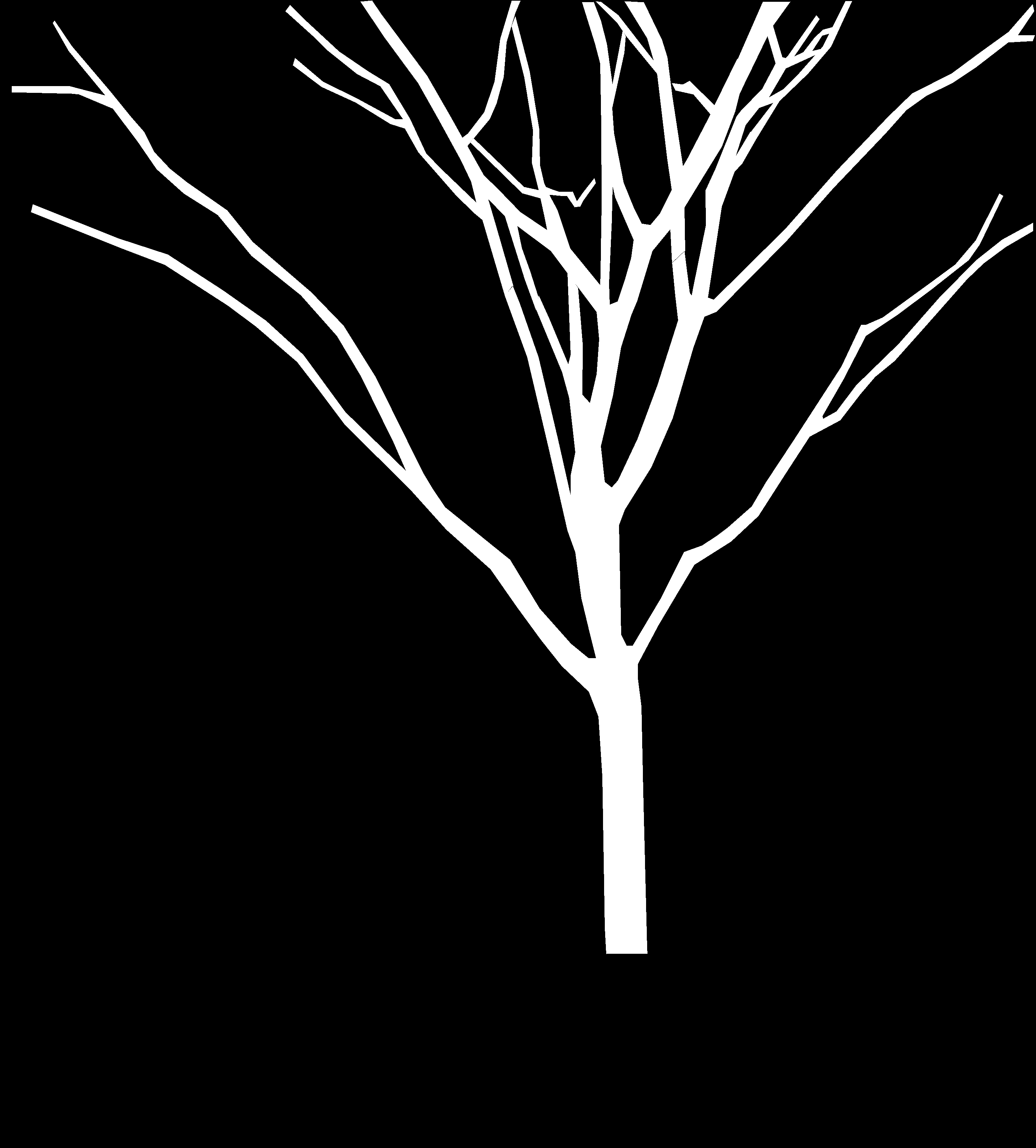}
            \caption{tree Mask}
        \end{subfigure}

        \vspace{3pt}
        \begin{subfigure}
            [t]{0.22\textwidth}
            \centering
            \includegraphics[width=\linewidth]{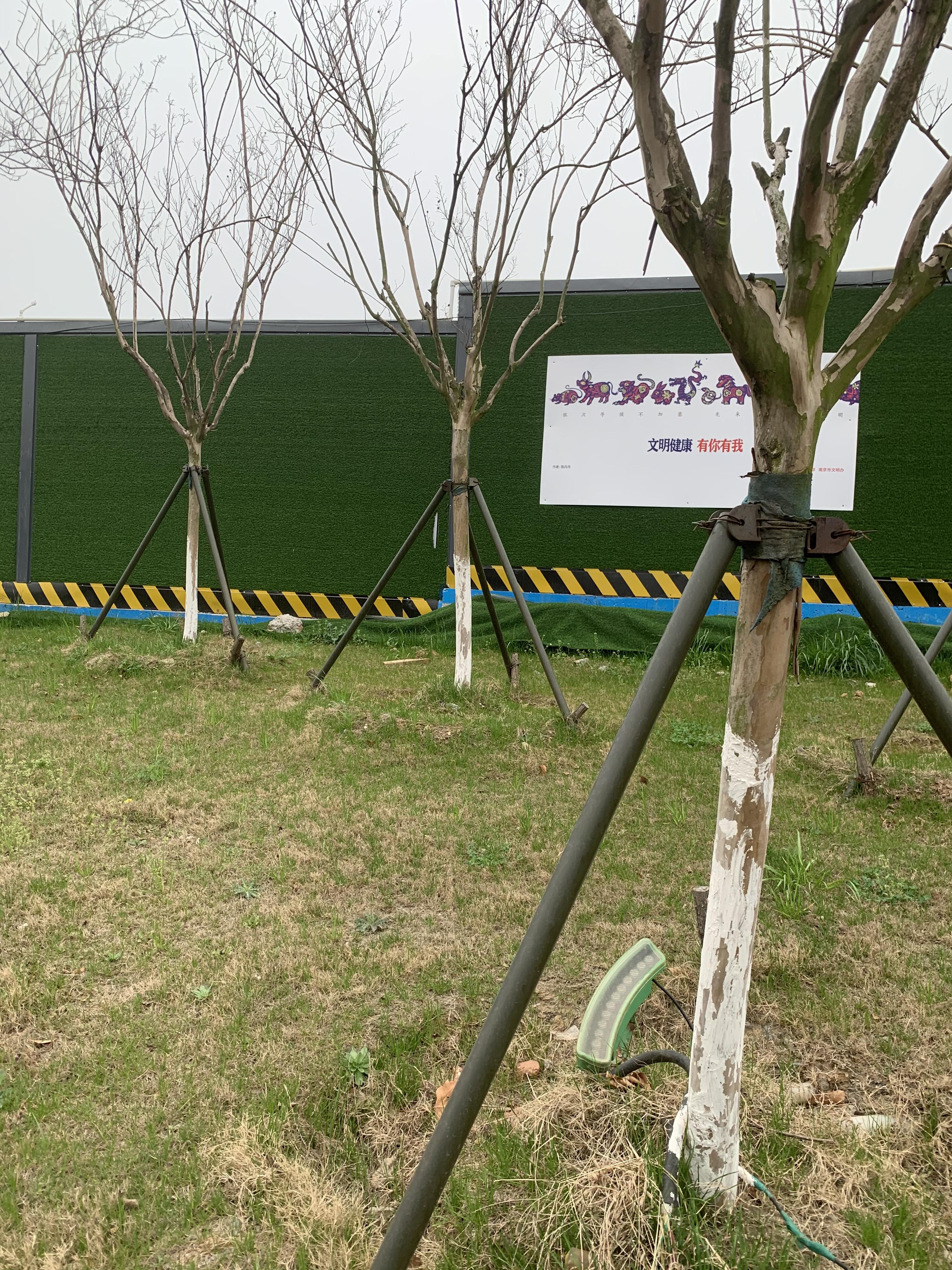}
            \caption{tree RGB image}
        \end{subfigure}\hfill
        \begin{subfigure}
            [t]{0.22\textwidth}
            \centering
            \includegraphics[width=\linewidth]{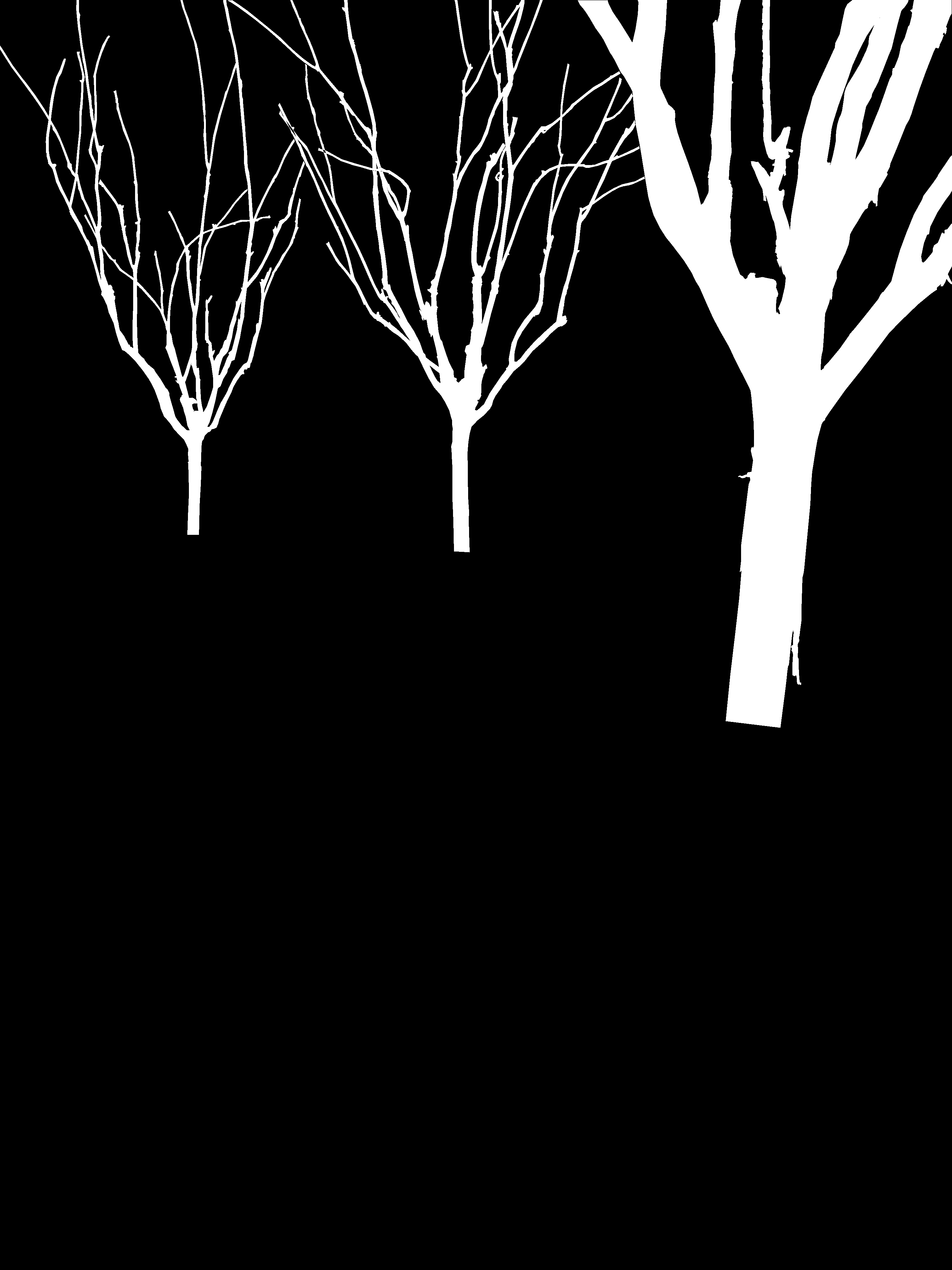}
            \caption{tree Mask}
        \end{subfigure}
        \caption{Example RGB images and ground-truth branch masks from the Urban
        Street Tree Dataset~\cite{kendric2023tree}.}
        \label{fig:dataset_examples}
    \end{figure}

    Accurate branch segmentation is important for vision-guided forestry systems.
    Tree branches present challenges that distinguish them from objects in standard
    computer vision datasets. These structures exhibit high aspect ratios (often
    exceeding 100:1), occupy small image areas (typically less than 3\% of total
    pixels), and form complex hierarchical patterns across multiple scales~\cite{zhang2021tree}
    (Fig.~\ref{fig:dataset_examples}). These characteristics differ from tasks in
    most existing datasets such as COCO~\cite{lin2014microsoft} or Cityscapes,
    which typically feature more balanced object distributions.

    Real-world deployment need to handle constraints beyond laboratory
    conditions. Autonomous forestry applications must handle variable lighting conditions,
    motion blur from UAV movement, diverse viewing angles, and complex
    backgrounds while maintaining performance on resource-constrained embedded
    platforms~\cite{steininger2025timbervision}. Safety requirements demand good
    precision—undetected branches pose operational risks, while false positives
    lead to unnecessary vegetation damage.

    Current research in precision forestry focuses primarily on canopy-level
    analysis for species classification and health monitoring~\cite{fassnacht2016review}.
    Individual branch segmentation remains relatively underexplored,
    particularly regarding embedded deployment requirements. Existing evaluation
    protocols rely predominantly on standard metrics that may inadequately capture
    the connectivity preservation and thin structure detection capabilities needed
    for structural analysis in forestry applications.

    This work aim to addresses these limitations through evaluation of deep learning
    architectures for branch segmentation at three input resolutions (256×256, 512×512,
    and 1024×1024). We assess segmentation heads and CNN/transformer backbones, introducing
    specialized evaluation metrics for thin structure analysis.

    The main contributions of this research are:

    \begin{itemize}
        \item Performance evaluation of segmentation heads and backbones across 256×256,
            512×512, and 1024×1024, analyzing accuracy–efficiency trade-offs (FLOPs)

        \item Two specialized evaluation metrics (TS-IoU, CPR) that complement
            IoU/Dice for thin-structure and topology assessment

        \item Comprehensive benchmark of 8 segmentation heads (U-Net, U-Net++,
            DeepLabV3/V3+, FPN, PSPNet, LinkNet, MAnet) with 14 backbones
            spanning CNN architectures (ResNet, MobileNet, DenseNet,
            EfficientNet, ConvNeXt) and vision transformers (ViT, Swin, LeViT,
            SegFormer), selected to represent diverse design paradigms and
            computational profiles

        \item Application-oriented guidance based on metric profiles and FLOPs for
            embedded selection
    \end{itemize}

    These contributions establish performance baselines and provide insights for
    computer vision applications in autonomous forestry.
    \vspace{-1 em}
    \section{Related Work}

    \subsection{Deep Learning for Semantic Segmentation}

    Deep learning based semantic segmentation methods have evolved through several
    architectural paradigms. Encoder-decoder networks, pioneered by U-Net~\cite{ronneberger2015u},
    employ symmetric architectures with skip connections to preserve spatial
    details during upsampling. Subsequent developments include U-Net++~\cite{zhou2018unet++},
    which introduces nested skip connections for improved feature propagation, Attention
    U-Net~\cite{oktay2018attention}, which incorporates attention mechanisms for
    selective feature enhancement, and U-Net3+~\cite{huang2020unet3+}, which implements
    full-scale feature aggregation.

    The DeepLab series~\cite{chen2018deeplabv3+} employ dilated convolution approaches
    to expand receptive fields without resolution loss. DeepLabV3+ combines Atrous
    Spatial Pyramid Pooling (ASPP) for multi-scale feature extraction with encoder-decoder
    refinement for boundary precision. PSPNet~\cite{zhao2017pyramid} utilizes
    spatial pyramid pooling for global context capture, though aggressive pooling
    may affect fine structure preservation. Recent transformer-based approaches
    like Mask2Former~\cite{cheng2022mask2former} show improved performance on
    standard benchmarks but require substantial computational resources (e.g., SegFormer
    MiT-B5 at 1024×1024: 281.6 GFLOPs vs.\ 37.7 GFLOPs for PSPNet) that may be
    challenging for embedded deployment.
    \vspace{-0.5 em}
    \subsection{Thin Structure Segmentation}

    Thin structure detection spans multiple application domains with varying
    technical challenges. Medical vessel segmentation shares structural similarities
    with branch detection, employing topology-preserving loss functions~\cite{shit2021cldice}
    and specialized attention mechanisms to maintain tubular connectivity. Infrastructure
    monitoring applications, including road network extraction and crack
    detection, address similar geometric challenges but typically involve more predictable
    patterns and controlled imaging conditions.

    Tree structure analysis remains limited in scope. Prior work~\cite{zhang2020pruning}
    demonstrated branch detection for fruit trees in controlled orchard
    environments, while subsequent research~\cite{wagner2017individual} focused on
    crown delineation for species identification rather than detailed structural
    analysis. These studies do not address challenges in autonomous forestry
    applications in uncontrolled environments.
    \vspace{-0.5 em}
    \subsection{Computer Vision in Forestry Applications}

    Current UAV-based forestry applications emphasize large-scale monitoring
    rather than detailed structural analysis. Established applications include forest
    health assessment, species classification using deep learning~\cite{natesan2019resnet},
    and biomass estimation. However, most systems operate at canopy level using imagery
    insufficient for individual branch analysis.

    Gaps exist between computer vision research and practical deployment requirements.
    Current evaluations predominantly use standard metrics (IoU/Dice) that may
    be inadequate for forestry-specific requirements such as thin structure
    detection, connectivity preservation, and structural integrity assessment. Additionally,
    computational constraints receive limited consideration despite being
    important for autonomous system deployment on resource-constrained UAV
    platforms.

    \section{Methodology}

    \subsection{Problem Formulation}

    We formulate tree branch segmentation as a binary pixel classification
    problem. Each pixel $p$ in input RGB image $\mathcal{I}\in \mathbb{R}^{H
    \times W \times 3}$ (where $H$ and $W$ denote image height and width, and 3 represents
    RGB color channels) is classified as either foreground (branch/trunk) or
    background. We evaluate at three resolutions:
    $H \times W \in \{256 \times 256, 512 \times 512, 1024 \times 1024\}$. The segmentation
    network learns a mapping $f_{\theta}: \mathbb{R}^{H \times W \times 3}\rightarrow
    [0,1]^{H \times W}$ parameterized by learnable weights $\theta$. The final
    binary segmentation mask $\mathcal{M}$ is obtained through thresholding: $\mathcal{M}
    (p) = \mathbb{I}[f_{\theta}(\mathcal{I})(p) > 0.5]$, where
    $\mathbb{I}[\cdot]$ denotes the indicator function.

    \subsection{Dataset}

    We utilize the Urban Street Tree Dataset~\cite{kendric2023tree}, containing
    1,485 high-resolution images (3024×4032 pixels) with pixel-level branch
    annotations. The dataset provides characteristics suitable for forestry applications:

    \begin{itemize}
        \item \textbf{High Resolution}: Native image resolution of 3024×4032 pixels
            provides details for fine branch structure analysis and supports multi-resolution
            evaluation.

        \item \textbf{Species Diversity}: Thirteen different tree species enhance
            model generalization capabilities.

        \item \textbf{Complex Backgrounds}: Natural urban environments with varying
            lighting conditions and occlusions mirror challenging conditions encountered
            in real-world forest deployments.

        \item \textbf{Binary Annotation}: Each pixel is labeled as either branch/trunk
            or background, providing ground truth for binary classification.
    \end{itemize}

    The dataset's species diversity is valuable for developing segmentation
    models that can generalize across different tree morphologies. This characteristic
    may be useful for future applications in Radiata pine forests, where models
    trained on diverse urban species could demonstrate adaptability to new
    coniferous structures.

    We employ a 60/20/20 train/validation/test split (891/297/297 images)
    stratified by tree species to ensure representative distribution across all
    evaluation phases.

    \subsection{Methods Implementation}

    We evaluate segmentation heads and backbones representative of current approaches,
    as summarized in Table~\ref{tab:methods}.
    \begin{table}[htb]
        \footnotesize
        \renewcommand{\arraystretch}{1.0} 
        \setlength{\tabcolsep}{4pt} 
        \centering
        \caption{Evaluated Segmentation Methods and Backbones}
        \label{tab:methods}
        \begin{tabularx}
            {\linewidth}{p{2.7cm} X} \toprule \textbf{Category} & \textbf{Methods}
            \\
            \midrule

            \multirow{2}{*}{\textbf{Segmentation Heads}} & U-Net~\cite{ronneberger2015u},
            U-Net++~\cite{zhou2018unet++}, DeepLabV3~\cite{chen2017deeplab}, DeepLabV3+~\cite{chen2018deeplabv3+}
            \\
            & FPN~\cite{lin2017fpn}, PSPNet~\cite{zhao2017pyramid}, LinkNet~\cite{chaurasia2017linknet},
            MAnet~\cite{ma2020manet} \\

            \midrule \multirow{2}{*}{\textbf{CNN Backbones}} & ResNet34/50~\cite{he2016resnet},
            MobileNetV2~\cite{sandler2018mobilenetv2}, DenseNet121~\cite{huang2017densenet}
            \\
            & EfficientNet-B0~\cite{tan2019efficientnet}, ConvNeXt-Base~\cite{liu2022convnext}
            \\

            \midrule \multirow{2}{*}{\textbf{Transformer Backbones}} & ViT-B/16~\cite{dosovitskiy2020vit},
            Swin-Base~\cite{liu2021swin}, LeViT-128s~\cite{graham2021levit} \\
            & MiT-B0--B5 (SegFormer)~\cite{xie2021segformer} \\

            \bottomrule
        \end{tabularx}
    \end{table}

    All methods are trained independently at each input resolution (256×256,
    512×512, 1024×1024) using AdamW optimizer (weight decay $10^{-4}$) with a
    base learning rate of $10^{-3}$, cosine annealing over 200 epochs, and early
    stopping based on validation loss. The loss function combines Binary Cross-Entropy
    (BCE) and Dice loss to address class imbalance:
    \begin{equation}
        \mathcal{L}= 0.3 \cdot \mathcal{L}_{BCE}+ 0.7 \cdot \mathcal{L}_{Dice}
    \end{equation}
    where:
    \begin{align}
        \mathcal{L}_{BCE}  & = -\frac{1}{N}\sum_{i=1}^{N}[y_{i}\log(\hat{y}_{i}) + (1-y_{i})\log(1-\hat{y}_{i})]                              \\
        \mathcal{L}_{Dice} & = 1 - \frac{2\sum_{i=1}^{N}y_{i}\hat{y}_{i}+ \epsilon}{\sum_{i=1}^{N}y_{i}+ \sum_{i=1}^{N}\hat{y}_{i}+ \epsilon}
    \end{align}

    The weighting emphasizes Dice loss for overlap quality while maintaining
    gradient stability through BCE inclusion.

    \subsection{Evaluation Metrics}

    Standard segmentation metrics may be insufficient for thin structure evaluation
    due to class imbalance and topology considerations. We employ conventional, specialized,
    and auxiliary metrics to assess performance:

    \textbf{Intersection over Union (IoU)}:
    \begin{equation}
        \text{IoU}= \frac{|Y \cap \hat{Y}|}{|Y \cup \hat{Y}|}
    \end{equation}

    \textbf{Dice Coefficient}:
    \begin{equation}
        \text{Dice}= \frac{2|Y \cap \hat{Y}|}{|Y| + |\hat{Y}|}
    \end{equation}
    where $Y$ represents ground truth and $\hat{Y}$ denotes predicted segmentation
    masks.

    \textbf{Thin Structure IoU (TS-IoU)}: Standard IoU may inadequately assess thin
    structure performance due to minimal pixel contribution. TS-IoU computes IoU
    specifically for branches with width less than 5 pixels:
    \begin{equation}
        \text{TS-IoU}= \frac{|Y_{thin}\cap \hat{Y}_{thin}|}{|Y_{thin}\cup \hat{Y}_{thin}|}
    \end{equation}
    where $Y_{thin}$ represents ground truth pixels belonging to morphologically
    identified thin branches. This metric addresses the challenge that thin
    branches, despite minimal image coverage, are important for structural
    integrity and safety assessment.

    \textbf{Connectivity Preservation Rate (CPR)}: Traditional pixel-wise
    metrics ignore topological relationships important for branch analysis. CPR measures
    the percentage of ground truth connected components adequately preserved in predictions:
    \begin{equation}
        \text{CPR}= \frac{1}{N_{cc}}\sum_{i=1}^{N_{cc}}\mathbb{I}[\text{IoU}(C_{i}
        , \hat{C}_{i}) > \tau]
    \end{equation}
    where $N_{cc}$ is the number of connected components in ground truth, $C_{i}$
    represents the $i$-th component, $\hat{C}_{i}$ is the corresponding
    prediction region (identified via maximum overlap), and $\tau = 0.5$ is the preservation
    threshold. This metric is useful for autonomous systems requiring branch
    connectivity understanding for safe operation.

    \subsubsection{Auxiliary Metrics}

    \textbf{Boundary-F1}: F1-score computed between contour maps of ground-truth
    and predictions. Contours are obtained via morphological gradient; matches are
    assessed within a small tolerance to account for annotation uncertainty.

    \textbf{Skeleton Similarity}: Dice coefficient computed between skeletonized
    ground-truth and predicted masks (morphological skeletonization), reflecting
    agreement on centerlines of thin structures.

    \subsubsection{Evaluation Framework Rationale}

    These complementary metrics address different aspects relevant for autonomous
    forestry:

    \begin{itemize}
        \item \textbf{IoU/Dice}: Overall segmentation quality for general performance
            assessment

        \item \textbf{TS-IoU}: Targeted evaluation of fine branch detection relevant
            to safety applications

        \item \textbf{CPR}: Topological correctness useful for path planning and
            structural analysis
    \end{itemize}

    This framework addresses both pixel-level accuracy and structural properties
    by integrating conventional metrics (IoU/Dice) with specialized metrics (TS-IoU
    and CPR), contrasting with standard benchmarks that emphasize overall accuracy
    without considering thin structure preservation.
    \vspace{-0.3 em}
    \section{Experimental Results}

    \subsection{Overall Performance across Resolutions}

    Tables~\ref{tab:main_results_256}--\ref{tab:main_results_1024} present performance
    comparisons at 256×256, 512×512, and 1024×1024 resolutions. Results are
    reported as validation/test pairs across all metrics.

    \begin{table*}
        [!t] \small
        \centering
        \caption{Performance Comparison at 256$\times$256 Resolution (Validation/Test)}
        \label{tab:main_results_256} \small
        \setlength{\tabcolsep}{2.5pt}
        \renewcommand{\arraystretch}{1.05}
        \begin{tabular}{p{1.6cm}p{1.8cm}|c|c|c|c|c|c|c|c|c}
            \toprule \textbf{Methods} & \textbf{Backbone}             & \textbf{IoU (\%)}             & \textbf{Dice (\%)}            & \textbf{Precision (\%)}       & \textbf{Recall (\%)}          & \textbf{TS-IoU (\%)}          & \textbf{CPR (\%)}    & \makecell{\textbf{Boundary-}\\\textbf{F1 (\%)}} & \makecell{\textbf{Skeleton}\\\textbf{Similarity}} & \makecell{\textbf{FLOPs}\\\textbf{(G)}} \\
            \midrule U-Net            & ResNet34                      & 65.1 / 65.0                   & 77.8 / 77.8                   & 79.6 / 80.3                   & 77.2 / 76.5                   & 67.4 / 67.9                   & 36.4 / 37.0          & 27.1 / 27.6                                     & 19.1 / 20.1                                       & 7.85                                    \\
            U-Net++                   & ResNet34                      & 69.1 / 69.6                   & 80.9 / 81.3                   & 82.5 / 83.8                   & 80.3 / 79.7                   & 71.6 / 72.5                   & 38.2 / 42.9          & 35.8 / 36.4                                     & 26.0 / 26.9                                       & 18.45                                   \\
            \footnotesize DeepLabV3   & ResNet34                      & 54.9 / 54.4                   & 69.3 / 68.9                   & 70.5 / 70.9                   & 69.6 / 68.6                   & 59.4 / 59.4                   & 34.2 / 35.2          & 11.4 / 11.7                                     & 7.6 / 7.5                                         & 27.351                                  \\
            \footnotesize DeepLabV3+  & ResNet34                      & 59.5 / 59.7                   & 73.2 / 73.5                   & 78.4 / 78.3                   & 70.0 / 70.5                   & 64.3 / 64.7                   & 35.6 / 35.7          & 18.8 / 19.3                                     & 12.1 / 12.4                                       & 7.925                                   \\
            FPN                       & ResNet34                      & 58.2 / 58.0                   & 72.1 / 72.2                   & 73.6 / 73.0                   & 72.1 / 72.5                   & 61.0 / 61.1                   & 33.5 / 32.4          & 15.8 / 16.0                                     & 11.4 / 11.7                                       & 6.875                                   \\
            PSPNet                    & ResNet34                      & 48.1 / 48.0                   & 62.8 / 62.8                   & 70.3 / 68.7                   & 59.0 / 59.9                   & 53.2 / 52.9                   & 31.8 / 31.5          & 9.0 / 9.0                                       & 5.5 / 5.5                                         & \textbf{2.359}                          \\
            LinkNet                   & ResNet34                      & 58.2 / 58.2                   & 72.5 / 72.3                   & 69.2 / 69.0                   & 78.1 / 77.7                   & 59.1 / 59.4                   & 32.3 / 31.7          & 17.6 / 17.9                                     & 13.1 / 13.1                                       & 5.474                                   \\
            MAnet                     & ResNet34                      & 63.2 / 63.1                   & 76.4 / 76.4                   & 78.3 / 78.8                   & 75.6 / 75.0                   & 65.0 / 65.7                   & 35.1 / 33.3          & 23.7 / 24.2                                     & 17.4 / 17.9                                       & 8.355                                   \\
            U-Net                     & ResNet50                      & 68.4 / 68.3                   & 80.3 / 80.3                   & 82.9 / 83.2                   & 78.6 / 78.5                   & 71.8 / 72.0                   & 37.6 / 41.5          & 34.2 / 34.8                                     & 24.3 / 24.8                                       & 10.707                                  \\
            U-Net                     & MobileNetV2                   & 62.4 / 63.0                   & 75.7 / 76.2                   & 77.6 / 77.6                   & 75.1 / 76.0                   & 63.7 / 65.0                   & 32.4 / 34.8          & 25.7 / 25.9                                     & 18.6 / 18.7                                       & 3.396                                   \\
            U-Net                     & DenseNet121                   & 67.2 / 67.9                   & 79.4 / 80.0                   & 82.6 / 83.2                   & 77.4 / 77.9                   & 69.8 / 70.7                   & 36.2 / 39.1          & 32.7 / 33.3                                     & 23.4 / 24.2                                       & 8.489                                   \\
            U-Net                     & \footnotesize EfficientNet-B0 & 66.2 / 66.9                   & 78.7 / 79.3                   & 79.9 / 80.4                   & 78.4 / 79.2                   & 67.9 / 69.0                   & 36.6 / 38.0          & 30.8 / 31.3                                     & 21.9 / 22.5                                       & 2.533                                   \\
            U-Net                     & \scriptsize ConvNeXt-Base     & 68.1 / 68.2                   & 80.2 / 80.2                   & 82.0 / 82.2                   & 79.2 / 79.1                   & 70.6 / 71.2                   & 37.5 / 41.8          & 33.7 / 34.2                                     & 23.9 / 24.4                                       & 20.836                                  \\
            U-Net                     & ViT-B/16                      & 65.1 / 64.8                   & 77.8 / 77.6                   & 80.2 / 80.2                   & 76.6 / 76.2                   & 67.5 / 67.4                   & 35.9 / 37.8          & 26.7 / 27.1                                     & 19.7 / 19.9                                       & 4.625                                   \\
            U-Net                     & Swin-Base                     & 65.1 / 64.8                   & 77.8 / 77.6                   & 80.4 / 80.4                   & 76.4 / 76.2                   & 67.7 / 67.5                   & 35.7 / 37.6          & 27.2 / 27.5                                     & 19.9 / 20.0                                       & 15.845                                  \\
            U-Net                     & LeViT-128s                    & 53.0 / 53.3                   & 67.7 / 68.0                   & 71.2 / 71.7                   & 67.5 / 67.5                   & 54.2 / 55.5                   & 31.6 / 34.6          & 19.6 / 19.5                                     & 14.2 / 14.2                                       & 2.974                                   \\
            U-Net                     & MiT-B0                        & 66.1 / 65.7                   & 78.6 / 78.3                   & 79.6 / 79.9                   & 78.3 / 77.5                   & 68.1 / 68.1                   & 36.2 / 36.0          & 25.1 / 25.9                                     & 18.4 / 18.8                                       & 2.932                                   \\
            U-Net                     & MiT-B1                        & 69.2 / 68.4                   & 80.9 / 80.3                   & 82.4 / 82.4                   & 80.0 / 79.0                   & 71.0 / 70.7                   & 36.6 / 39.8          & 30.7 / 31.0                                     & 22.0 / 22.3                                       & 4.882                                   \\
            U-Net                     & MiT-B2                        & 71.1 / 70.4                   & 82.3 / 81.7                   & 84.7 / 84.4                   & 80.5 / 79.9                   & 74.0 / 73.7                   & 42.2 / 42.6          & 34.0 / 34.0                                     & 24.4 / 24.6                                       & 6.887                                   \\
            U-Net                     & MiT-B3                        & 71.3 / 70.6                   & 82.3 / 81.9                   & \textbf{85.5} / \textbf{85.7} & 80.0 / 79.2                   & 74.5 / 74.6                   & 42.1 / 41.6          & 35.3 / 35.5                                     & 25.1 / 25.1                                       & 10.549                                  \\
            U-Net                     & MiT-B4                        & \textbf{73.8} / \textbf{72.6} & \textbf{84.2} / \textbf{83.4} & 84.8 / 83.9                   & \textbf{84.2} / \textbf{83.3} & \textbf{76.1} / \textbf{75.0} & \textbf{43.6} / 43.6 & \textbf{39.2} / \textbf{39.1}                   & \textbf{28.7} / \textbf{28.4}                     & 14.065                                  \\
            U-Net                     & MiT-B5                        & 72.5 / 71.8                   & 83.3 / 82.7                   & 84.6 / 84.7                   & 82.4 / 81.4                   & 74.9 / 74.9                   & 41.5 / \textbf{44.2} & 37.2 / 37.3                                     & 26.8 / 26.9                                       & 17.647                                  \\
            \bottomrule
        \end{tabular}
    \end{table*}

    \begin{table*}
        [!t] \small
        \centering
        \caption{Performance Comparison at 512$\times$512 Resolution (Validation/Test)}
        \label{tab:main_results_512} \small
        \setlength{\tabcolsep}{2.5pt}
        \renewcommand{\arraystretch}{1.05}
        \begin{tabular}{p{1.6cm}p{1.8cm}|c|c|c|c|c|c|c|c|c}
            \toprule \textbf{Methods} & \textbf{Backbone}             & \textbf{IoU (\%)}             & \textbf{Dice (\%)}            & \textbf{Precision (\%)}       & \textbf{Recall (\%)}          & \textbf{TS-IoU (\%)}          & \textbf{CPR (\%)}    & \makecell{\textbf{Boundary-}\\\textbf{F1 (\%)}} & \makecell{\textbf{Skeleton}\\\textbf{Similarity}} & \makecell{\textbf{FLOPs}\\\textbf{(G)}} \\
            \midrule U-Net            & ResNet34                      & 77.9 / 77.4                   & 87.1 / 86.7                   & 87.7 / 87.6                   & 86.9 / 86.4                   & 79.8 / 79.3                   & 24.4 / 24.3          & 34.1 / 34.1                                     & 26.6 / 26.6                                       & 31.40                                   \\
            U-Net++                   & ResNet34                      & 80.2 / 79.5                   & 88.6 / 88.2                   & 89.5 / 89.3                   & 88.1 / 87.6                   & 82.2 / 81.5                   & 29.0 / 26.9          & 38.9 / 38.9                                     & 29.9 / 30.1                                       & 73.80                                   \\
            \footnotesize DeepLabV3+  & ResNet34                      & 71.6 / 71.4                   & 82.6 / 82.6                   & 84.6 / 84.2                   & 81.7 / 81.7                   & 74.6 / 74.5                   & 22.1 / 24.0          & 20.6 / 21.4                                     & 14.6 / 15.3                                       & 31.70                                   \\
            \footnotesize DeepLabV3   & ResNet34                      & 69.2 / 68.6                   & 80.9 / 80.5                   & 80.3 / 79.8                   & 82.0 / 81.7                   & 72.0 / 71.4                   & 26.2 / 27.6          & 16.1 / 16.4                                     & 11.6 / 11.8                                       & 109.40                                  \\
            FPN                       & ResNet34                      & 71.7 / 71.4                   & 82.7 / 82.6                   & 84.6 / 84.7                   & 81.6 / 81.2                   & 74.6 / 74.4                   & 21.7 / 23.4          & 20.7 / 21.2                                     & 15.1 / 15.8                                       & 27.50                                   \\
            PSPNet                    & ResNet34                      & 64.2 / 63.7                   & 76.9 / 76.5                   & 80.7 / 80.1                   & 74.5 / 74.2                   & 68.2 / 67.7                   & 22.9 / 24.1          & 12.5 / 12.6                                     & 8.3 / 8.4                                         & \textbf{9.43}                           \\
            LinkNet                   & ResNet34                      & 74.5 / 74.0                   & 84.8 / 84.4                   & 86.1 / 86.1                   & 84.1 / 83.5                   & 76.6 / 76.1                   & 12.3 / 12.3          & 28.1 / 28.2                                     & 21.1 / 21.3                                       & 21.90                                   \\
            MAnet                     & ResNet34                      & 77.0 / 76.8                   & 86.4 / 86.3                   & 86.9 / 86.6                   & 86.7 / 86.7                   & 78.6 / 78.4                   & 25.8 / 26.7          & 33.1 / 33.4                                     & 25.5 / 25.8                                       & 33.42                                   \\
            U-Net                     & ResNet50                      & 79.0 / 78.5                   & 87.8 / 87.5                   & 88.4 / 88.2                   & 87.6 / 87.2                   & 81.1 / 80.5                   & 28.4 / 28.9          & 36.7 / 37.2                                     & 27.6 / 28.2                                       & 42.83                                   \\
            U-Net                     & MobileNetV2                   & 76.2 / 75.4                   & 85.8 / 85.3                   & 86.7 / 86.3                   & 85.8 / 85.4                   & 78.1 / 77.1                   & 19.5 / 18.3          & 32.9 / 32.8                                     & 24.8 / 24.9                                       & 13.58                                   \\
            U-Net                     & DenseNet121                   & 78.5 / 78.5                   & 87.5 / 87.5                   & 88.2 / 88.3                   & 87.2 / 87.3                   & 80.2 / 80.1                   & 21.5 / 23.3          & 36.5 / 36.8                                     & 28.0 / 28.7                                       & 33.96                                   \\
            U-Net                     & \footnotesize EfficientNet-B0 & 78.4 / 78.3                   & 87.4 / 87.3                   & 88.1 / 88.4                   & 87.4 / 87.0                   & 80.4 / 80.6                   & 18.9 / 18.6          & 36.3 / 36.6                                     & 27.3 / 27.8                                       & 10.13                                   \\
            U-Net                     & \scriptsize ConvNeXt-Base     & 79.0 / 78.7                   & 87.8 / 87.6                   & 89.0 / 88.8                   & 87.0 / 87.1                   & 81.2 / 80.8                   & 26.5 / 30.5          & 36.6 / 37.1                                     & 27.7 / 28.0                                       & 56.26                                   \\
            U-Net                     & ViT-B/16                      & 77.8 / 77.0                   & 87.0 / 86.4                   & 87.8 / 87.5                   & 86.7 / 85.9                   & 79.7 / 79.1                   & 26.4 / 25.2          & 34.1 / 34.1                                     & 26.4 / 26.6                                       & 27.55                                   \\
            U-Net                     & Swin-Base                     & 77.9 / 77.3                   & 87.1 / 86.7                   & 87.5 / 87.2                   & 87.2 / 86.7                   & 79.6 / 79.2                   & 25.2 / 26.1          & 34.4 / 34.4                                     & 26.8 / 26.8                                       & 56.26                                   \\
            U-Net                     & LeViT-128s                    & 66.9 / 67.3                   & 79.0 / 79.3                   & 80.2 / 80.9                   & 79.4 / 79.3                   & 68.8 / 69.5                   & 16.1 / 16.0          & 25.3 / 25.6                                     & 18.9 / 19.0                                       & 11.89                                   \\
            U-Net                     & MiT-B0                        & 80.1 / 79.4                   & 88.6 / 88.1                   & 89.2 / 89.0                   & 88.3 / 87.6                   & 82.0 / 81.2                   & 26.4 / 29.3          & 35.6 / 35.9                                     & 27.0 / 27.5                                       & 11.73                                   \\
            U-Net                     & MiT-B1                        & 81.5 / 80.8                   & 89.4 / 89.0                   & 90.3 / 90.4                   & 88.9 / 88.1                   & 83.3 / 82.8                   & 32.8 / 31.1          & 38.5 / 38.7                                     & 29.8 / 30.1                                       & 19.53                                   \\
            U-Net                     & MiT-B2                        & 82.9 / 82.3                   & 90.3 / 89.9                   & 90.7 / 90.3                   & 90.2 / 90.0                   & 85.0 / 84.0                   & 33.2 / 35.4          & 42.1 / 42.7                                     & 32.0 / 32.7                                       & 27.55                                   \\
            U-Net                     & MiT-B3                        & 83.3 / 82.6                   & 90.6 / 90.1                   & \textbf{91.7} / \textbf{91.3} & 89.8 / 89.2                   & 85.3 / 84.4                   & 33.5 / 33.6          & 42.3 / 42.9                                     & 32.5 / 32.9                                       & 42.20                                   \\
            U-Net                     & MiT-B4                        & \textbf{83.8} / \textbf{83.1} & \textbf{90.9} / \textbf{90.4} & 91.4 / 91.1                   & 90.6 / 90.1                   & \textbf{85.9} / \textbf{85.0} & 35.7 / \textbf{36.3} & \textbf{44.0} / \textbf{44.6}                   & 33.6 / \textbf{34.0}                              & 56.26                                   \\
            U-Net                     & MiT-B5                        & 83.6 / 82.9                   & 90.8 / 90.3                   & 90.9 / 90.5                   & \textbf{90.9} / \textbf{90.5} & 85.7 / 84.8                   & \textbf{38.7} / 34.3 & 43.8 / 44.3                                     & \textbf{33.7} / 33.6                              & 70.59                                   \\
            \bottomrule
        \end{tabular}
    \end{table*}

    \begin{table*}
        [!t] \small
        \centering
        \caption{Performance Comparison at 1024$\times$1024 Resolution (Validation/Test)}
        \label{tab:main_results_1024} \small
        \setlength{\tabcolsep}{2.5pt}
        \renewcommand{\arraystretch}{1.05}
        \begin{tabular}{p{1.6cm}p{1.8cm}|c|c|c|c|c|c|c|c|c}
            \toprule \textbf{Methods} & \textbf{Backbone}             & \textbf{IoU (\%)}    & \textbf{Dice (\%)}   & \textbf{Precision (\%)}       & \textbf{Recall (\%)}          & \textbf{TS-IoU (\%)} & \textbf{CPR (\%)}             & \makecell{\textbf{Boundary-}\\\textbf{F1 (\%)}} & \makecell{\textbf{Skeleton}\\\textbf{Similarity}} & \makecell{\textbf{FLOPs}\\\textbf{(G)}} \\
            \midrule U-Net            & ResNet34                      & 83.9 / 83.3          & 91.0 / 90.6          & 91.5 / 90.9                   & 90.9 / 90.8                   & 84.9 / 84.2          & 13.0 / 14.6                   & 29.6 / 29.9                                     & 24.4 / 24.5                                       & 125.596                                 \\
            U-Net++                   & ResNet34                      & 86.3 / 85.8          & 92.4 / 92.1          & 92.3 / 92.3                   & 92.8 / 92.3                   & 87.1 / 86.5          & 17.3 / 18.4                   & \textbf{35.8} / \textbf{36.8}                   & \textbf{28.8} / \textbf{29.4}                     & 295.197                                 \\
            \footnotesize DeepLabV3   & ResNet34                      & 80.8 / 80.2          & 88.9 / 88.6          & 89.2 / 88.8                   & 89.0 / 88.7                   & 81.9 / 81.4          & 11.1 / 12.0                   & 20.3 / 20.8                                     & 15.3 / 15.9                                       & 437.616                                 \\
            \footnotesize DeepLabV3+  & ResNet34                      & 83.3 / 82.6          & 90.7 / 90.2          & 90.7 / 90.5                   & 91.0 / 90.2                   & 84.1 / 83.5          & 11.8 / 13.0                   & 25.3 / 26.1                                     & 19.4 / 19.8                                       & 126.801                                 \\
            FPN                       & ResNet34                      & 80.6 / 80.3          & 88.8 / 88.7          & 89.7 / 89.1                   & 88.5 / 88.5                   & 81.6 / 81.3          & 10.1 / 11.9                   & 20.9 / 21.6                                     & 16.5 / 17.1                                       & 110.005                                 \\
            PSPNet                    & ResNet34                      & 75.5 / 75.3          & 85.4 / 85.2          & 87.4 / 86.8                   & 84.0 / 84.1                   & 76.7 / 76.5          & 5.3 / 6.6                     & 14.7 / 15.3                                     & 11.0 / 11.4                                       & \textbf{37.735}                         \\
            LinkNet                   & ResNet34                      & 82.8 / 82.5          & 90.3 / 90.1          & 90.7 / 90.5                   & 90.2 / 90.0                   & 83.8 / 83.4          & 2.9 / 3.7                     & 27.0 / 27.5                                     & 21.8 / 22.2                                       & 87.589                                  \\
            MAnet                     & ResNet34                      & 84.8 / 84.1          & 91.5 / 91.0          & 91.6 / 91.1                   & 91.7 / 91.5                   & 85.7 / 84.9          & 11.3 / 13.5                   & 31.1 / 31.5                                     & 25.3 / 25.6                                       & 133.689                                 \\
            U-Net                     & ResNet50                      & 84.6 / 83.9          & 91.4 / 90.9          & 91.8 / 91.3                   & 91.4 / 91.0                   & 85.6 / 84.8          & 13.7 / 14.3                   & 31.3 / 31.6                                     & 25.4 / 25.7                                       & 171.310                                 \\
            U-Net                     & MobileNetV2                   & 81.7 / 81.2          & 89.6 / 89.2          & 90.7 / 89.7                   & 88.9 / 89.4                   & 82.7 / 81.9          & 6.9 / 7.7                     & 27.6 / 27.7                                     & 22.5 / 22.5                                       & 54.339                                  \\
            U-Net                     & DenseNet121                   & 84.8 / 84.1          & 91.5 / 91.0          & 92.1 / 91.8                   & 91.2 / 90.8                   & 85.7 / 84.8          & 10.3 / 13.4                   & 30.8 / 31.6                                     & 25.4 / 26.0                                       & 135.825                                 \\
            U-Net                     & \footnotesize EfficientNet-B0 & 83.8 / 83.3          & 90.9 / 90.6          & 91.2 / 90.9                   & 91.0 / 90.7                   & 84.8 / 84.1          & 7.2 / 8.6                     & 29.7 / 30.1                                     & 24.0 / 24.1                                       & 40.530                                  \\
            U-Net                     & \scriptsize ConvNeXt-Base     & 84.6 / 84.2          & 91.4 / 91.2          & 91.1 / 91.2                   & 92.0 / 91.5                   & 85.5 / 85.0          & 10.7 / 14.5                   & 31.2 / 31.7                                     & 25.2 / 25.6                                       & 61.305                                  \\
            U-Net                     & ViT-B/16                      & 84.3 / 83.7          & 91.2 / 90.8          & 91.3 / 90.8                   & 91.5 / 91.2                   & 85.2 / 84.5          & 13.5 / 14.5                   & 29.8 / 30.2                                     & 24.6 / 24.8                                       & 53.708                                  \\
            U-Net                     & Swin-Base                     & 84.1 / 83.4          & 91.1 / 90.7          & 91.5 / 91.1                   & 91.0 / 90.6                   & 85.0 / 84.3          & 11.7 / 14.4                   & 29.6 / 29.9                                     & 24.5 / 24.6                                       & 73.948                                  \\
            U-Net                     & LeViT-128s                    & 73.9 / 74.6          & 84.2 / 84.7          & 85.6 / 86.6                   & 83.7 / 83.6                   & 74.9 / 75.5          & 2.2 / 2.7                     & 21.5 / 21.8                                     & 17.1 / 17.2                                       & 47.576                                  \\
            U-Net                     & MiT-B0                        & 86.0 / 85.3          & 92.2 / 91.8          & 92.2 / 92.3                   & 92.4 / 91.6                   & 86.8 / 86.1          & 16.6 / 17.8                   & 32.0 / 32.3                                     & 26.0 / 26.1                                       & 46.814                                  \\
            U-Net                     & MiT-B1                        & 86.6 / 86.2          & 92.6 / 92.3          & 92.8 / 92.7                   & 92.6 / 92.3                   & 87.4 / 86.9          & 16.3 / 17.9                   & 33.8 / 34.5                                     & 27.4 / 28.0                                       & 77.915                                  \\
            U-Net                     & MiT-B2                        & 87.0 / 86.6          & 92.8 / 92.6          & \textbf{93.3} / \textbf{93.3} & 92.5 / 92.2                   & 87.8 / 87.2          & 20.6 / 20.3                   & 34.1 / 34.9                                     & 27.7 / 28.3                                       & 109.874                                 \\
            U-Net                     & MiT-B3                        & \textbf{87.3} / 86.8 & \textbf{93.1} / 92.7 & \textbf{93.4} / 93.1          & 92.9 / 92.6                   & \textbf{88.1} / 87.4 & 25.4 / 25.1                   & 35.0 / 35.7                                     & 28.5 / 29.0                                       & 168.322                                 \\
            U-Net                     & MiT-B4                        & 87.2 / \textbf{87.1} & 92.9 / \textbf{92.9} & 93.1 / 92.9                   & \textbf{93.0} / \textbf{93.2} & 87.9 / \textbf{87.7} & 24.8 / 24.0                   & 35.4 / 36.2                                     & 28.7 / \textbf{29.4}                              & 224.405                                 \\
            U-Net                     & MiT-B5                        & 87.2 / \textbf{87.1} & 93.0 / \textbf{92.9} & 93.2 / 93.2                   & \textbf{93.0} / 92.9          & 88.0 / \textbf{87.7} & \textbf{25.5} / \textbf{25.8} & 35.3 / 36.2                                     & 28.6 / 29.3                                       & 281.599                                 \\
            \bottomrule
        \end{tabular}
    \end{table*}

    \subsection{Performance Analysis}

    \paragraph{256×256 (Table~\ref{tab:main_results_256}).}
    U-Net+MiT-B4 performs well overall, leading in IoU, Dice, Recall, TS-IoU,
    Boundary-F1, and Skeleton Similarity on validation and maintaining high performance
    on test data. Precision is highest with U-Net+MiT-B3 on both validation and
    test, whereas CPR is highest on test with U-Net+MiT-B5. PSPNet provides the
    lowest compute cost (2.36 GFLOPs) with reduced accuracy.

    \paragraph{512×512 (Table~\ref{tab:main_results_512}).}
    U-Net+MiT-B4 achieves good IoU and Dice performance on validation and test,
    along with strong TS-IoU and boundary quality metrics. U-Net+MiT-B3 yields high
    precision on validation and test (91.7\%/91.3\%), while U-Net+MiT-B5
    achieves high recall on validation and test (90.9\%/90.5\%) and good CPR on
    validation (38.7\%). Test CPR is highest with MiT-B4 (36.3\%). PSPNet again
    offers the lowest FLOPs (9.43G) with reduced accuracy.

    \paragraph{1024×1024 (Table~\ref{tab:main_results_1024}).}
    U-Net+MiT-B3 provides good validation IoU/Dice and precision; U-Net+MiT-B4/\,B5
    achieve strong test IoU/Dice performance, with recall being high with MiT-B4.
    Boundary quality (Boundary-F1) is best with U-Net++ on both validation and
    test, and Skeleton Similarity is highest on test with MiT-B4. FLOPs increase
    substantially at this scale; PSPNet remains the most efficient option (37.74G)
    but with reduced accuracy.

    \paragraph{Cross-resolution trends.}
    Increasing input resolution improves absolute scores across architectures.
    Transformer backbones (MiT) scale favorably compared to CNNs, maintaining good
    IoU/Dice and thin-structure metrics (TS-IoU, Boundary-F1, Skeleton
    Similarity). Precision vs. recall trade-offs persist: MiT-B3 favors
    precision, while MiT-B4/B5 emphasize recall and boundary fidelity. CPR
    values are lower than pixel metrics, reflecting the difficulty of preserving
    topology; they improve with stronger backbones and higher resolution.

    \subsection{Notes on Efficiency}

    The table reports theoretical FLOPs to characterize computational cost.
    While device-specific throughput (FPS), memory, and power depend on
    implementation and hardware, the FLOPs trends indicate accuracy–efficiency trade-offs
    useful for embedded selection.
    \vspace{-0.5 em}
    \section{Discussions}
    \vspace{-0.3 em}
    \subsection{Application-Oriented Guidance}

    Based on the multi-resolution evaluation in Tables~\ref{tab:main_results_256}--\ref{tab:main_results_1024},
    we provide method selection guidance for different application requirements:

    \textbf{When overall accuracy is critical}: U-Net + MiT-B4 achieves the highest
    IoU/Dice scores at 256×256 (73.8\%/84.2\%) and 512×512 (83.8\%/90.9\%),
    making it suitable for general-purpose segmentation where balanced
    performance across all metrics is required.

    \textbf{When precision matters most}: U-Net + MiT-B3 delivers superior
    precision at 512×512 (91.7\% validation), reducing false positives—critical
    for safety-critical applications where erroneous branch detection could
    cause navigation errors.

    \textbf{When computational resources are limited}: Different efficiency
    levels require different approaches: (1) PSPNet provides the lowest FLOPs (2.36/9.43/37.74G
    across resolutions) but sacrifices 25.7/19.6/11.8 percentage points in IoU; (2)
    EfficientNet-B0 and MiT-B0 backbones offer better accuracy–efficiency trade-offs
    than PSPNet at comparable computational costs; (3) LinkNet and FPN present
    intermediate options balancing accuracy and efficiency.
    \vspace{-0.3 em}
    \subsection{Thin Structure Detection}

    TS-IoU and skeleton similarity metrics reveal that fine branch detection
    remains challenging across all architectures. At 256×256, the best-performing
    model achieves 76\% TS-IoU, indicating room for improvement in sub-pixel
    structure preservation. Potential enhancement strategies include: (i) Multi-scale
    training with resolution augmentation. (ii) Topology-aware loss functions incorporating
    connectivity constraints. (iii) Post-processing refinement using
    morphological operations. (iv) Ensemble methods combining predictions across
    multiple scales.
    \vspace{-0.3 em}
    \subsection{Deployment Considerations}

    For autonomous forestry applications, we recommend a three-stage strategy: (i)
    conservative detection thresholds optimized for high recall to minimize missed
    branches, (ii) temporal consistency validation across video sequences to
    filter false positives, and (iii) human oversight integration for high-risk scenarios.
    The reported per-frame metrics provide baselines that should be validated
    under real-world operational conditions including variable lighting, motion blur,
    and diverse environmental contexts.
    \vspace{-0.5 em}
    \section{Conclusions}

    This work evaluated deep learning architectures for tree branch segmentation
    across three resolutions (256×256, 512×512, 1024×1024) to support UAV-based
    autonomous forestry operations. Using the Urban Street Tree Dataset, we
    assessed both standard metrics (IoU, Dice) and specialized measures (TS-IoU,
    CPR, boundary-F1, skeleton similarity) to capture thin-structure
    preservation and topology maintenance critical for forestry applications. Our
    evaluation of 22 configurations reveals that U-Net with MiT-B4 backbone delivers
    strong performance at lower resolutions, while transformer backbones
    consistently outperform CNNs across all resolutions despite higher
    computational costs. At 1024×1024, MiT-B3/B4/B5 variants demonstrate
    competitive performance depending on specific precision-recall and boundary
    quality requirements. These findings establish multi-resolution benchmarks that
    inform accuracy–efficiency trade-offs for embedded deployment in autonomous forestry
    systems. Future work should address resolution-agnostic training approaches,
    topology-aware loss functions, and comprehensive real-world validation to
    bridge the gap between laboratory performance and operational deployment in
    UAV-based forestry applications.
    \vspace{-1.5em}
    \section{Acknowledgement}
    This work was supported in part by the MBIE Endeavor Research Programme UOCX2104,
    the MBIE Data Science SSIF Fund under contract RTVU1914, and Marsden Fund of
    New Zealand Government under Contracts VUW2115.

    \bibliographystyle{IEEEtran}
    
\end{document}